\documentclass[runningheads,a4paper]{llncs}
\usepackage{hyperref}
\usepackage{amssymb}
\usepackage{graphicx,epstopdf,epsfig,subcaption,amsmath}
\usepackage{tikz}
\usepackage{setspace}
\usepackage{color}
\usepackage{hhline}
\usepackage{url}
\usepackage{multirow}
\usepackage[font=small]{caption}
\usepackage{array,booktabs}
\usepackage{color}
\usepackage{ctable}

\urldef{\mailsa}\path|{alfred.hofmann, ursula.barth, ingrid.haas, frank.holzwarth,|
\urldef{\mailsb}\path|anna.kramer, leonie.kunz, christine.reiss, nicole.sator,|
\urldef{\mailsc}\path|erika.siebert-cole, peter.strasser, lncs}@springer.com|

\usepackage{floatrow}
\usepackage{subcaption}
\usepackage{float}
\newfloatcommand{capbtabbox}{table}[][\FBwidth]

\newsavebox\CBox
\def\textBF#1{\sbox\CBox{#1}\resizebox{\wd\CBox}{\ht\CBox}{\textbf{#1}}}
\usepackage{algorithm}
\usepackage{algorithmicx}
\usepackage{algpseudocode}
\usepackage{wrapfig,lipsum}

\begin{document}
\mainmatter  

\makeatletter
\def\@normalsize{\@setsize\normalsize{10pt}\xpt\@xpt
\abovedisplayskip 10pt plus2pt minus5pt\belowdisplayskip
\abovedisplayskip \abovedisplayshortskip \z@
plus3pt\belowdisplayshortskip 6pt plus3pt
minus3pt\let\@listi\@listI}
\def\subsize{\@setsize\subsize{12pt}\xipt\@xipt}
\def\section{\@startsection {section}{1}{\z@}{1.0ex plus
1ex minus .2ex}{.2ex plus .2ex}{\large\bf}}
\def\subsection{\@startsection {subsection}{2}{\z@}{.2ex
plus 1ex} {.2ex plus .2ex}{\subsize\bf}} \makeatother

\newcommand{\Section}[1]{\section{\hskip -1em.~~#1}}
\newcommand{\SubSection}[1]{\subsection{\hskip -1em.~~#1}}
\def\@listI{%
 \leftmargin\leftmargini
 \partopsep 0pt
 \parsep 0pt
 \topsep 0pt
 \itemsep pt
 \relax
} \long\def\@makecaption#1#2{
 \vskip -5pt
 \setbox\@tempboxa\hbox{\small{#1\,:\,#2}}
  \ifdim \wd\@tempboxa >\hsize \unhbox\@tempboxa\par \else
  \hbox to\hsize{\hfil\box\@tempboxa\hfil}
\fi \vskip -0.2cm}

\jot=0pt \abovedisplayskip=3pt \belowdisplayskip=3pt
\abovedisplayshortskip=0pt \belowdisplayshortskip=0pt

\newcommand{\etal}{\textit{et al}.}
\newcommand{\ie}{\textit{i}.\textit{e}.}
\newcommand{\eg}{\textit{e}.\textit{g}.}
\newcommand{\aph}[1]{\textcolor{red}{#1}}

\title{Accurate Weakly-Supervised Deep Lesion Segmentation using Large-Scale Clinical Annotations: Slice-Propagated 3D Mask Generation from 2D RECIST}

\author{Jinzheng Cai\inst{1,2}\thanks{indicates equal contribution}, Youbao Tang\inst{1}\protect\footnotemark[1], Le Lu\inst{1}, Adam P. Harrison\inst{1}, Ke Yan\inst{1}, \\
	Jing Xiao\inst{3}, Lin Yang\inst{2}, and Ronald M. Summers\inst{1} \\
	\institute{National Institutes of Health, Bethesda, MD, 20892, USA
		\and University of Florida, Gainesville, FL, 32611, USA
		\and Ping An Insurance (Group) Company of China, Ltd., Shenzhen, 510852, PRC
		\email{jimmycai@ufl.edu}, \email{\{youbao.tang}, \email{le.lu}, \email{adam.harrison}, \email{ke.yan\}@nih.gov}, 
		\email{xiaojing661@pingan.com.cn}, \email{lin.yang@bme.ufl.edu},  \email{rms@nih.gov}
	}
}

\titlerunning{Accurate Weakly-Supervised Deep Lesion Segmentation}
\authorrunning{J. Cai, Y. Tang~\etal{}}

%
%
%


%
%

\toctitle{Lecture Notes in Computer Science}
\tocauthor{Authors' Instructions}


\maketitle
\vspace{-9mm}
\begin{abstract}
Volumetric lesion segmentation from computed tomography (CT) images is a powerful means to precisely assess multiple time-point lesion/tumor changes. 
However, because manual 3D segmentation is prohibitively time consuming, current practices rely on an imprecise surrogate called response evaluation criteria in solid tumors (RECIST). 
Despite their coarseness, RECIST markers are commonly found in current hospital picture and archiving systems (PACS), meaning they can provide a potentially powerful, yet extraordinarily challenging, source of weak supervision for full 3D segmentation. 
Toward this end, we introduce a convolutional neural network (CNN) based weakly supervised slice-propagated segmentation (WSSS) method to 1) generate the initial lesion segmentation on the axial RECIST-slice; 2) learn the data distribution on RECIST-slices; 3) extrapolate to segment the whole lesion slice by slice to finally obtain a volumetric segmentation. 
To validate the proposed method, we first test its performance on a fully annotated lymph node dataset, where WSSS performs comparably to its fully supervised counterparts. 
We then test on a comprehensive lesion dataset with $32,735$ RECIST marks, where we report a mean Dice score of $92\%$ on RECIST-marked slices and $76\%$ on the entire 3D volumes. 
\end{abstract}

\section{Introduction}
Given the prevailing clinical adoption of the response evaluation criteria in solid tumors (RECIST) for cancer patient monitoring~\cite{eisenhauer_2009_recist,ybtang_recist}, many modern hospitals' picture archiving and communication systems (PACS) store tremendous amounts of lesion diameter measurements linked to computed tomography (CT) images. In this paper, we tackle the challenging problem of leveraging existing RECIST diameters to produce fully volumetric lesion segmentations in 3D. From any input  CT image with the RECIST diameters, we first segment the lesion on the RECIST-marked image (RECIST-slice) in a weakly supervised manner, followed by generalizing the process into other successive slices to obtain the lesion's full volume segmentation. 

\par \qquad  Inspired by related work~\cite{j_dai_iccv15_boxsup,khoreva_cvpr17_simpledoesit,d_lin_cvpr16_scribblesup,g_papandreou_iccv15_wssl} of weakly supervised segmentation in computer vision, we design our lesion segmentation in an iteratively slice-wise propagated fashion. More specifically, with the bookmarked long and short diameters on the RECIST-slice, we initialize the segmentation using unsupervised learning methods, \eg, GrabCut~\cite{rother_2004_grabcut}. Afterward, we iteratively refine the segmentation using a supervised convolutional neural network (CNN), which can accurately segment the lesion on RECIST-slices. Importantly, the resulting CNN model, trained from all RECIST-slices, can capture the appearance of lesions in CT slices. Thus, the model is capable of detecting lesion regions from images other than the RECIST-slices. With more slices segmented, more image data can be extracted and used to further fine-tune the model. As such, the proposed weakly supervised segmentation model is a slice-wise label-map propagation process, from the RECIST-slice to the whole lesion volume. Therefore, we leverage a large amount of retrospective (yet clinically annotated) imaging data to automatically achieve the final 3D lesion volume measurement and segmentation. 

\par \qquad  To compare the proposed weakly supervised slice-propagated segmentation (WSSS) against a fully-supervised upper performance limit, we first validate on a publicly-available lymph node (LN) dataset~\cite{roth_2014_lymphnode}, consisting of $984$ LNs with full pixel-wise annotations. After demonstrating comparable performance to fully-supervised approaches, we then evaluate WSSS on the DeepLesion dataset~\cite{jzcai_deeplesion,yan_2017_deep-lesion}, achieving mean DICE scores of $92\%$ and $76\%$ on the RECIST-slices and lesion volumes, respectively. 

\section{Method}
In the DeepLesion dataset~\cite{jzcai_deeplesion,yan_2017_deep-lesion}, each CT volume contains an axial slice marked with RECIST diameters that represent the longest lesion axis and its perpendicular counterpart. RECIST diameters can act as a means of weakly supervised training data. Thus, we leverage weakly supervised principles to learn a CNN model using CT slices with no extra pixel-wise manual annotations. Formally, we denote elements in DeepLesion as {\small $\{(V^i,R^i)\}$} for {\small $i \in \{1,\ldots,N\}$}, where {\small $N$} is the number of lesions, {\small $V^i$} is the CT volume of interest, and {\small $R^i$} is the corresponding RECIST diameter. To create the 2D training data for the segmentation model, the RECIST-slice and label pairs, {\small $X^i$} and {\small $Y^i$}, respectively, must be generated, and {\small $X^i=V_{r}^i$} is simply the RECIST-slice, \ie, the axial slice at index $r$ that contains {\small $R$}. For notational clarity, we drop the superscript {\small $i$} for the remainder of this discussion. 

\subsection{Initial RECIST-Slice Segmentation} \label{sec:init-mask}
We adopt GrabCut~\cite{rother_2004_grabcut} to produce the initial lesion segmentation on RECIST-slices. GrabCut is initialized with image foreground and background seeds, {\small $Y^{s}$}, and produces a segmentation using iterative energy minimization. The resulting mask is calculated to minimize an objective energy function conditioned on the input CT image and seeds: 
\begin{equation} \label{eq:gc}
Y = \underset{\tilde{Y}}{\operatorname{arg\,min}}~E_{gc} (\tilde{Y},Y^{s},X) \textrm{,}
\end{equation}
where we follow the original definition of the energy function {\small $E_{gc}$} in~\cite{rother_2004_grabcut}. 

\par \qquad  Given the fact that the quality of GrabCut's initialization will largely affect the final result, we propose to use the spatial prior information, provided by {\small $R$}, to compute  high quality initial seeds, {\small $Y^{s}=S(R)$}, where {\small $S(R)$} produces four categories: regions of background (BG), foreground (FG), \emph{probable} background (PBG), and \emph{probable} foreground (PFG). More specifically, if the lesion bounding box tightly around the RECIST axes is {\small $[w,h]$}, a {\small $[2w,2h]$} region of interest (ROI) is cropped from the RECIST-slice. The outer 50\% of the ROI is assigned to BG whereas 10\% of the image region, obtained from a dilation around {\small $R$} is assigned to FG. The remaining 40\% is divided between PFG and PBG based on the distances to FG and BG. Fig.~\ref{fig:overview} visually depicts the training mask generation process (see the ``RECIST to Mask'' part). We use FG and BG as GrabCut seed regions, leaving the rest as regions where the initial mask is estimated. 

\subsection{RECIST-Slice Segmentation} \label{sec:cnn}
We represent our CNN model as a mapping function {\small $\hat{Y} = f(X;\theta)$}, where {\small $\theta$} represents the model parameters. Our goal is to minimize the differences between {\small $\hat{Y}$} and the imperfect GrabCut mask {\small $Y$}, which contains 3 groups, namely the RECIST pixel indices {\small $\mathcal{R}$}, the estimated lesion (foreground) pixel indices {\small $\mathcal{F}$}, and the estimated background pixel indices set {\small $\mathcal{B}$}. Formally, the indices sets are defined to satisfy the constraints as {\small $Y = Y_\mathcal{R} \cup Y_\mathcal{F} \cup Y_\mathcal{B}$}, and {\small $\mathcal{R} \cap \mathcal{F} = \mathcal{R} \cap \mathcal{B} = \mathcal{F} \cap \mathcal{B} = \emptyset$}. Thus, we define CNN's training objective containing 3 loss parts as, 
\begin{eqnarray} \label{eq:convnet}
L &=& L_{\mathcal{R}} + \alpha L_{\mathcal{F}} + \beta L_{\mathcal{B}} \textrm{,} \\
&=& \frac{1}{|\mathcal{R}|} \sum_{i \in \mathcal{R}} -\log{\hat{y}_i} + \alpha \frac{1}{|\mathcal{F}|} \sum_{i \in \mathcal{F}} -\log{\hat{y}_i} + \beta \frac{1}{|\mathcal{B}|} \sum_{i \in \mathcal{B}} -\log{(1 - \hat{y}_i)} \textrm{,}
\end{eqnarray}
where {\small $\hat{y}_i$} is the {\small $i^{th}$} pixel in {\small $\hat{Y}$}, {\small $|\cdot|$} represents the set cardinality and {\small $\alpha,\beta$} are positive weights to balance the losses. Empirically, we set {\small $\alpha$}, and {\small $\beta$} to small values at the start of model training when {\small $\mathcal{F},\mathcal{B}$} regions are estimated with low confidence. Afterwards, we set {\small $\alpha$}, and {\small $\beta$} to larger values, \eg, 1, when training converges. 

\subsection{Weakly Supervised Slice-Propagated Segmentation} \label{sec:3d-wsss}
To obtain volumetric measurements, we follow a similar strategy as with the RECIST-slices, except in this slice-propagated case, we must infer {\small $R$} for off-RECIST-slices and also incorporate inference results {\small $\hat{Y}$} from the CNN model. These two priors are used together for slice-propagated CNN training.

\begin{figure*}[t!]
	\begin{center}
		
%
		\includegraphics[width=.9\linewidth, height=3.2cm, page=6, trim={0cm 6.3cm 2.6cm 0cm}, clip]{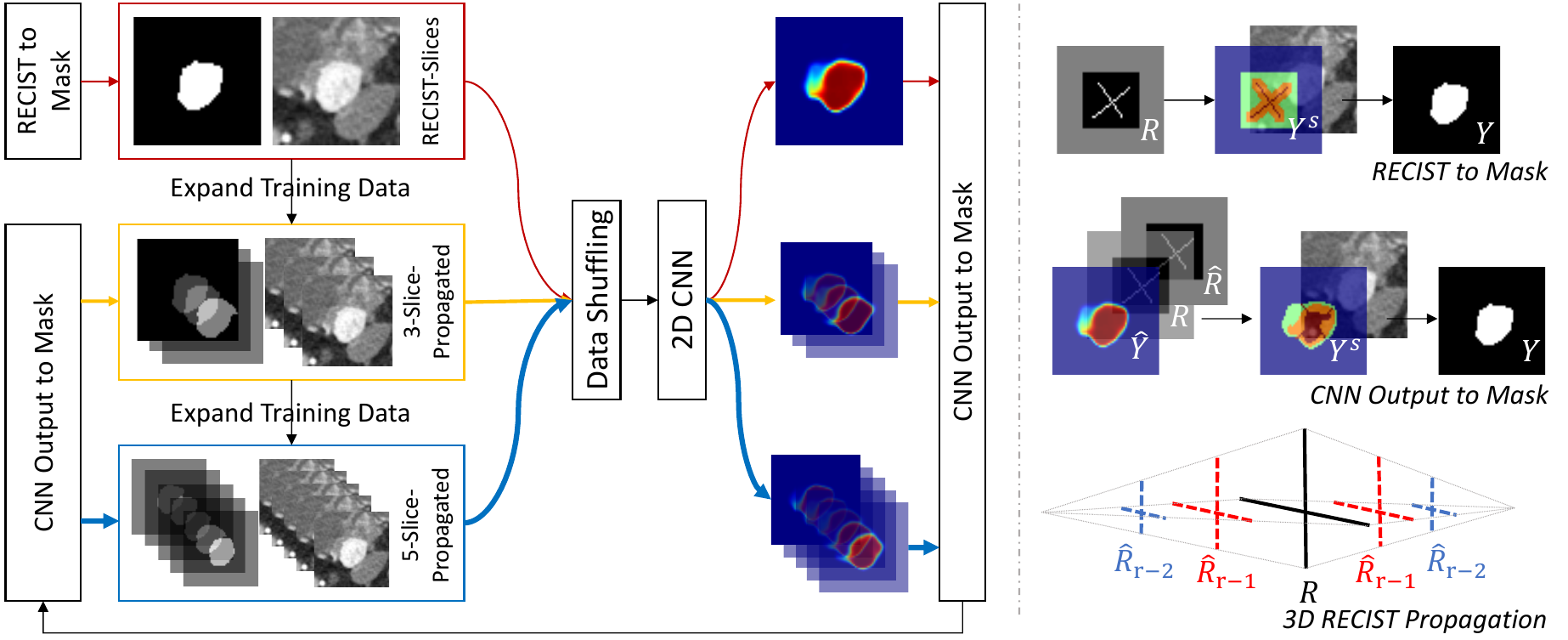} 
	\end{center}
	\vspace{-4mm}
	\caption{
	Overview of the proposed method.
	{\bf Right:} we use CNN outputs to gradually generate extra training data for lesion segmentation. 
	Arrows colored in red, orange, and blue indicate slice-propagated training at its $1^{st}$, $2^{nd}$, and $3^{rd}$ steps, respectively. 
	{\bf Left:} regions colored with red, orange, green, and blue inside the initial segmentation mask $Y$ present FG, PFG, PBG, and BG, respectively.  
	Best viewed in color. \vspace{-7mm}}
	\label{fig:overview}
\end{figure*}

\textbf{RECIST Propagation:} A simple way to generate off-RECIST-slice diameters {\small $\hat{R}$} is to take advantage of the fact that RECIST-slice {\small $R$} lies on the maximal cross-sectional area of the lesion. The rate of reduction of off-RECIST-slice endpoints is then calculated by their relative offset distance to the RECIST-slice. Propagated RECIST endpoints are then projected from the actual RECIST endpoints by the Pythagorean theorem using physical Euclidean distance. The ``3D RECIST Propagation'' part in Fig.~\ref{fig:overview} depicts the propagation across CT slices. Given the actual RECIST on the {\small $r^{th}$} slice, {\small $\hat{R}_{r-1}$} and {\small $\hat{R}_{r-2}$} are the estimated RECISTs on the first and second off-RECIST-slices, respectively.

\textbf{Off-RECIST-Slice Segmentation:} For slice {\small $r$}, offset from the RECIST-slice, we update the seed generation function from  Sec.~\ref{sec:init-mask} to now take both the inference from the RECIST-slice trained CNN, {\small $\hat{Y}$}, and the estimated RECIST, {\small $\hat{R}$}: {\small$Y^{s}=S(\hat{Y},\hat{R},R)$}. More specifically, {\small $\hat{Y}$} is first binarized by adjusting the threshold so that it covers at least 50\% of {\small $R$}'s pixels. Regions in {\small $\hat{Y}$} that associate with high foreground probability values, \ie, {\small $>0.8$}, and overlap with {\small $\hat{R}$} will be set as FG together with {\small $\hat{R}$}. Similarly, regions with high background probabilities and that have no overlap with {\small $\hat{R}$} will be assigned as BG. The remaining pixels are left as uncertain using the same distance criteria as in the 2D mask generation case and fed into GrabCut for lesion segmentation. In the limited cases where the CNN fails to detect any foreground regions, we fall back to seed generation in Sec.~\ref{sec:init-mask}, except we use {\small $\hat{R}$} as input. The GrabCut mask is then generated using Equation~\eqref{eq:gc} as before. This procedure is also visually depicted in Fig.~\ref{fig:overview} (see the ``CNN Output to Mask'' part).

\textbf{Slice-Propagated CNN Training:} To generate lesion segmentations in all CT slices from 2D RECIST annotations, we train the CNN model in a slice-propagated manner. The CNN first learns lesion appearances based on the RECIST-slices. After the model converges, we then apply this CNN model to slices {\small $[V_{r-1},V_{r+1}]$} from the entire training set to compute initial predicted probability maps {\small $[\hat{Y}_{r-1},\hat{Y}_{r+1}]$}. Given these probability maps, we create initial lesion segmentations {\small $[Y_{r-1}, Y_{r+1}]$} using GrabCut and the seed generation explained above. These segmentations are employed as training labels for the CNN model on the {\small $[V_{r-1}, V_{r+1}]$} slices, ultimately producing the finally updated segmentations {\small $[\hat{Y}_{r-1},\hat{Y}_{r+1}]$} once the model converges. As this procedure proceeds iteratively, we can gradually obtain the converged lesion segmentation result across CT slices, and then stack the slice-wise segmentations {\small $[\ldots,\hat{Y}_{r-1},\hat{Y}_r,\hat{Y}_{r+1},\ldots]$} to produce a volumetric segmentation. We visually depict this process in Fig.~\ref{fig:overview} from RECIST-slice to 5 successive slices. 

\section{Results} 
\textbf{Datasets:} The DeepLesion dataset~\cite{jzcai_deeplesion,yan_2017_deep-lesion} is composed of $32,735$ bookmarked CT lesion instances (with RECIST measurements) from $10,594$ studies of $4,459$ patients. Lesions have been categorized into $8$ subtypes: lung, mediastinum (MD), liver, soft-tissue (ST), abdomen (AB), kidney, pelvis, and bone. For quantitative evaluation, we segmented $1,000$ testing lesion RECIST-slices manually. Out of these $1000$, $200$ lesions ($\sim3,500$ annotated slices) are fully segmented in 3D as well. Additionally, we also employ the lymph node (LN) dataset\footnote{\url{https://wiki.cancerimagingarchive.net/display/Public/CT+Lymph+Nodes}} \cite{roth_2014_lymphnode}, which consists of $176$ CT scans with complete pixel-wise annotations. Enlarged LN is a lesion subtype and producing accurate segmentation is quite challenging even with fully supervised learning~\cite{n_isabella_miccai2016_lymphnode}. Importantly, the LN dataset can be used to evaluate our WSSS method against an upper-performance limit, by comparing results with a fully supervised approach~\cite{n_isabella_miccai2016_lymphnode}. 

{\bf Pre-processing:} For the LN dataset, annotation masks are converted into RECIST diameters by measuring its major and minor axes. For robustness, up to $20\%$ random noise is injected into the RECIST diameter lengths to mimic the uncertainty of manual annotation by radiologists. For both datasets, based on the location of RECIST bookmarks, CT ROIs are cropped at two times the extent of the lesion's longest diameters so that sufficient visual context is preserved. The dynamic range of each lesion ROI is then intensity-windowed properly using the CT windowing meta-information in \cite{yan_2017_deep-lesion}. The LN dataset is separated at the patient level, using a split of $80\%$ and $20\%$ for training and testing, respectively. For the DeepLesion \cite{yan_2017_deep-lesion} dataset, we randomly select $28,000$ lesions for training.

{\bf Evaluation:} The mean DICE similarity coefficient (mDICE) and the pixel-wise precision and recall are used to evaluate the quantitative segmentation accuracy.

\subsection{Initial RECIST-Slice Segmentation} \label{exp:recist-seg}
We denote the seed and GrabCut generation approach in Sec.~\ref{sec:init-mask} as GrabCut-R. In addition, we test two seed generation alternatives, which are based on a tight bounding box (bbox) matching the extent of the lesion RECIST marks with 25\% padding. The first alternative (GrabCut) sets the areas inside and outside the bbox as BG and PFG, respectively. The second alternative (GrabCut$^i$) sets the central 20\% bbox region as FG, regions outside the bbox as BG, and the rest as PFG. This is similar to the setting of bbox$^i$ in~\cite{khoreva_cvpr17_simpledoesit}. We also test the densely connected conditional random fields (DCRF)~\cite{krah_2012_fully-crf}, using bbox$^i$ as the unary potentials and intensities to compute pairwise potentials~\cite{krah_2012_fully-crf}. As the DCRF was moderately sensitive to parameter variations, we report the best configuration we found in Table~\ref{tab:mask-init}. Finally, we also report results when we directly use the RECIST diameters, but dilated to 20\% of bbox area, to generate {\small $Y$}. We denote this approach RECIST-D, which produces the best precision, but at the cost of very low recall. However, as can be seen in Table~\ref{tab:mask-init}, GrabCut-R significantly outperforms all alternatives on both of the LN and the DeepLesion datasets, demonstrating the validity of our mask initialization process. 

\begin{table}[t!]
\begin{center}
\caption{Performance in generating $Y$, the initial RECIST-slice segmentation. Mean DICE scores are reported with standard deviation for methods that defined in Sec.~\ref{exp:recist-seg}.} \vspace{-3mm}
\label{tab:mask-init}
{\scriptsize
\begin{tabular}{>{~}m{2cm}>{\centering}m{1.5cm} >{\centering}m{1.5cm} >{\centering}m{1.5cm} >{\centering}m{0.1cm} >{\centering}m{1.5cm} >{\centering}m{1.5cm} >{\centering\arraybackslash}m{1.5cm}}
\toprule
& \multicolumn{3}{c}{Lymph Node} & & \multicolumn{3}{c}{DeepLesion (on RECIST-Slice)} \\ \cmidrule{2-4} \cmidrule{6-8}
Method & Recall & Precision & mDICE & & Recall & Precision & mDICE \\ \midrule
RECIST-D   & 0.35$\pm$0.09  		& \textBF{0.99$\pm$0.05}& 0.51$\pm$0.09 & 		   & 0.39$\pm$0.13  		& \textBF{0.92$\pm$0.14} & 0.53$\pm$0.14 \\
DCRF       & 0.29$\pm$0.20  		& 0.98$\pm$0.05  		& 0.41$\pm$0.21 & 		   & 0.72$\pm$0.26  		& 0.90$\pm$0.15  		 & 0.77$\pm$0.20 \\
GrabCut    & 0.10$\pm$0.25  		& 0.32$\pm$0.37  		& 0.11$\pm$0.26 & 		   & 0.62$\pm$0.46  		& 0.68$\pm$0.44  		 & 0.62$\pm$0.46 \\
GrabCut$^i$& 0.53$\pm$0.24  		& 0.92$\pm$0.10  		& 0.63$\pm$0.17 & 		   & 0.94$\pm$0.11  		& 0.81$\pm$0.16  		 & 0.86$\pm$0.11 \\
GrabCut-R  & \textBF{0.83$\pm$0.11} & 0.86$\pm$0.11  		& \textBF{0.83$\pm$0.06} & & \textBF{0.94$\pm$0.10} & 0.89$\pm$0.10  		 & \textBF{0.91$\pm$0.08} \\
\bottomrule
\end{tabular}
}
\vspace{-6mm}
\end{center}
\end{table}


\subsection{RECIST-Slice Segmentation} \label{exp:cnn}
We use holistically nested networks (HNNs)~\cite{xie_2015_hnn} as our baseline CNN model, which has been adapted successfully for lymph node~\cite{n_isabella_miccai2016_lymphnode}, pancreas~\cite{j_cai_miccai17_pancreas}, and lung segmentation~\cite{Harrison_2017}. In all experiments, deep learning is implemented in Tensorflow~\cite{tensorflow2015-whitepaper} and Tensorpack~\footnote{\url{https://github.com/ppwwyyxx/tensorpack}} with pre-trained models. The initial learning rate is {\small $5\times10^{-5}$}, dropping to {\small $1\times10^{-5}$} when the model training-validation plot plateaus. Given the results of {\small $Y$}, \ie, $>$90\% mDICE, we simply set the balance weights in Equation~\eqref{eq:convnet} as $\alpha=\beta=1$.

\par \qquad  Following Sec.~\ref{exp:recist-seg}, we select three ways to generate training masks on the RECIST-slice: the RECIST-D, GrabCut-R and the fully annotated ground truth (GT). As Table~\ref{tab:2d-convnet} demonstrates, on the LN dataset\cite{roth_2014_lymphnode}, HNNs trained using masks {\small $Y$} generated from RECIST-D, GrabCut-R, and GT achieve 61\%, 70\%, and 71\% mDICE scores, respectively. This observation demonstrates the robustness and effectiveness of using GrabCut-R labels, which only performs slightly worse than using the GT. On the DeepLesion~\cite{yan_2017_deep-lesion} testset of $1,000$ annotated RECIST-slices, HNN trained on GrabCut-R outperforms the deep model learned from RECIST-D by a margin of 25\% in mean DICE (90.6\% versus 64.4\%). GrabCut post-processing, denoted with the suffix ``-GC'', further improves the results from 90.6\% to 91.5\%. We also aim to demonstrate that our weakly supervised approach, trained on a large quantity of ``imperfectly-labeled'' object masks, can outperform fully-supervised models trained on fewer data. To do this, we separated the $1,000$ annotated testing images into five folds and report the mean DICE scores using fully-supervised HNN~\cite{xie_2015_hnn} and UNet~\cite{ronneberger_2015_unet} models on this smaller dataset. Impressively, the 90.6\% DICE score of the weakly supervised approach considerably outperforms the fully supervised HNN and UNet mDICE of 83.7\% and 72.8\%, respectively. Coupled with an approach like ours, this demonstrates the potential in exploiting large-scale, but ``imperfectly-labeled'', datasets. 

\subsection{Weakly Supervised Slice-Propagated Segmentation} \label{exp:wsss}
\begin{figure}[t!]
	\centering
	\mbox{%
		\begin{minipage}[b]{0.55\linewidth}
			\centering	
			\subcaption{~}	
			\label{fig:subA}
			\vspace{-1mm}	
			\includegraphics[width=\linewidth, height=2.8cm,trim={1cm 0cm 1.5cm 0cm}, clip]{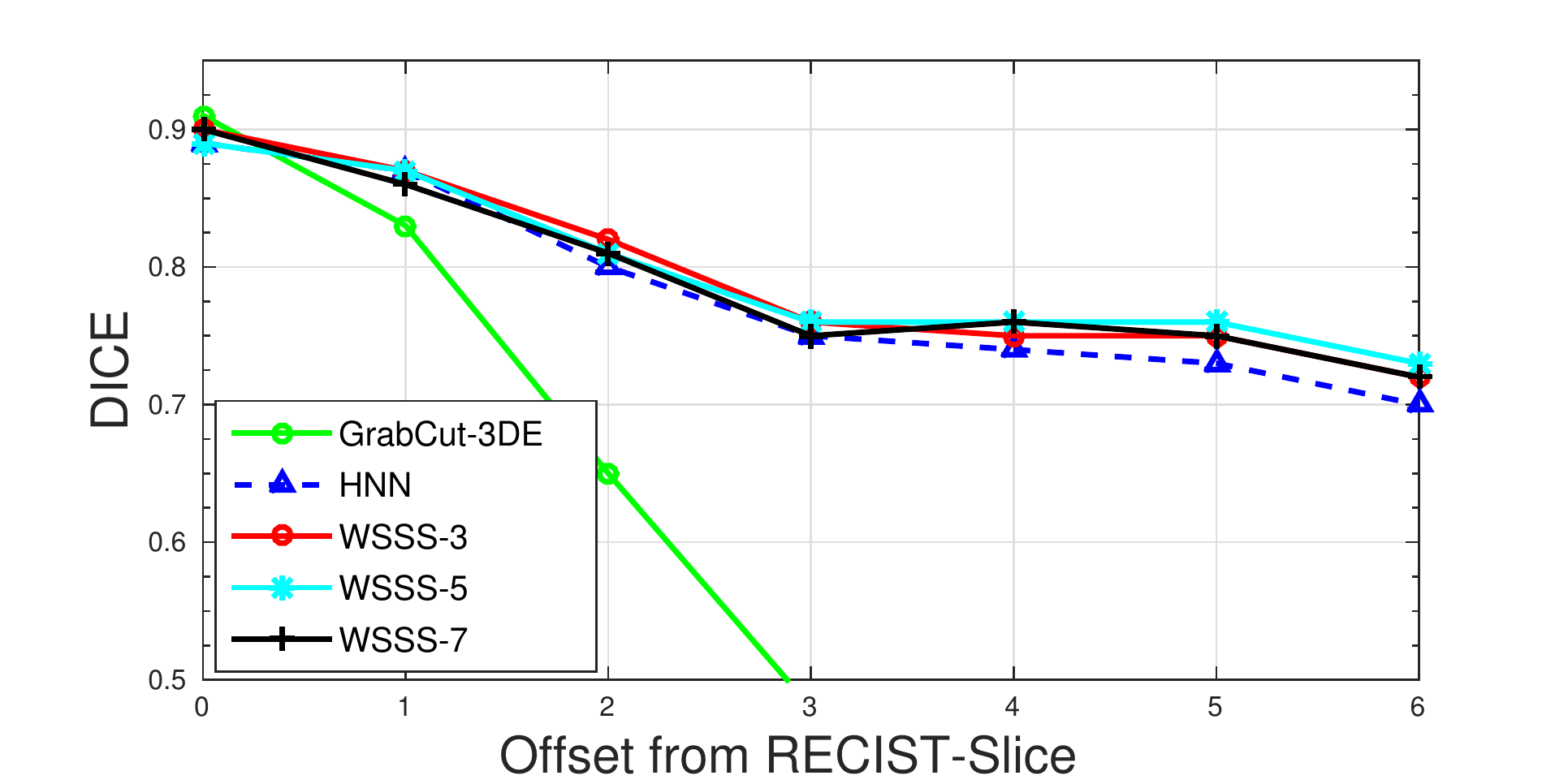}    
		\end{minipage}
		\begin{minipage}[b]{0.35\linewidth}
			\centering			
			\subcaption{~}
			\label{fig:subB}
			\vspace{-1mm}
			\includegraphics[width=\linewidth,height=2.8cm,trim={0cm 0cm 1cm 0cm}, clip]{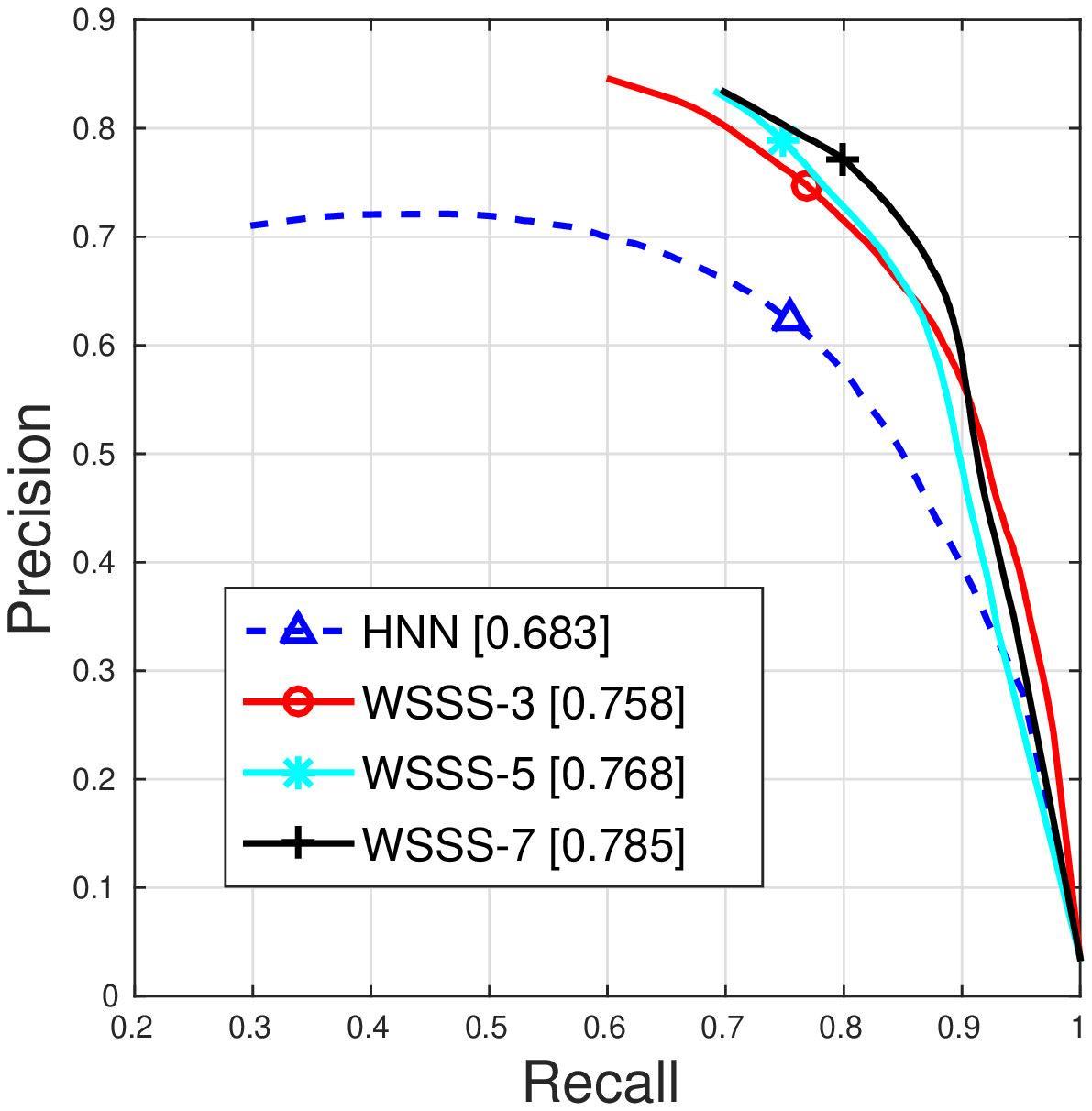}
		\end{minipage}		
}
	\vspace{-2mm}
	\caption{WSSS on DeepLesion. 
	(a) depicts mean Dice scores on \emph{2D slices} as a function of offsets with respect to the RECIST-slice. 
	(b) depicts \emph{volumetric} precision-recall curves. \vspace{-6mm}}
	\label{fig:curves}
\end{figure}

\begin{figure*}[t]
\begin{center}
\setlength{\fboxsep}{0pt}
 \hspace{0.05\linewidth} \hfill
 \fbox{\includegraphics[width=0.1\linewidth, height=0.1\linewidth]{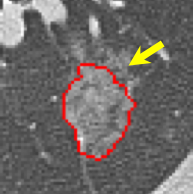}} \hfill
 \fbox{\includegraphics[width=0.1\linewidth, height=0.1\linewidth]{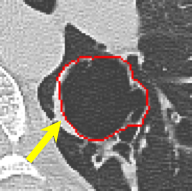}} \hfill
 \fbox{\includegraphics[width=0.1\linewidth, height=0.1\linewidth]{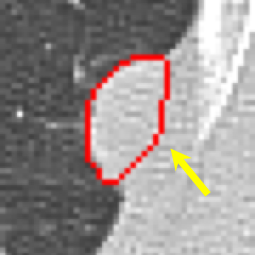}} \hfill
 \fbox{\includegraphics[width=0.1\linewidth, height=0.1\linewidth]{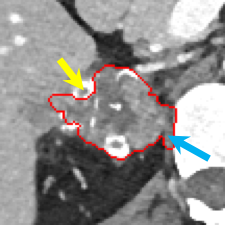}} \hfill
 \fbox{\includegraphics[width=0.1\linewidth, height=0.1\linewidth]{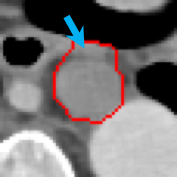}} \hfill
 \fbox{\includegraphics[width=0.1\linewidth, height=0.1\linewidth]{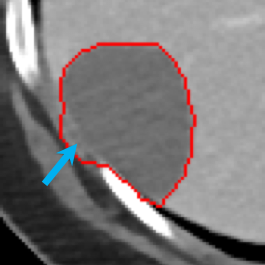}} \hfill
 \fbox{\includegraphics[width=0.1\linewidth, height=0.1\linewidth]{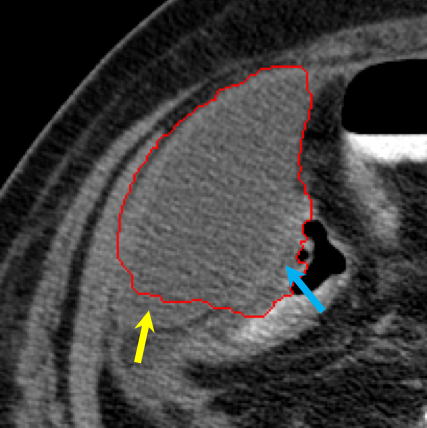}}  \hfill
 \fbox{\includegraphics[width=0.1\linewidth, height=0.1\linewidth]{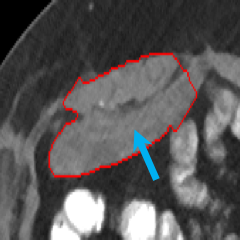}} 
 \hspace{0.05\linewidth} \hfill\\[1mm]
 \hspace{0.05\linewidth} \hfill
 \fbox{\includegraphics[width=0.1\linewidth, height=0.1\linewidth]{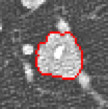}} \hfill
 \fbox{\includegraphics[width=0.1\linewidth, height=0.1\linewidth]{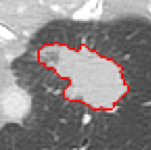}} \hfill
 \fbox{\includegraphics[width=0.1\linewidth, height=0.1\linewidth]{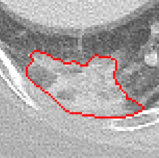}} \hfill
 \fbox{\includegraphics[width=0.1\linewidth, height=0.1\linewidth]{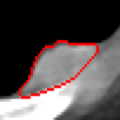}} \hfill
 \fbox{\includegraphics[width=0.1\linewidth, height=0.1\linewidth]{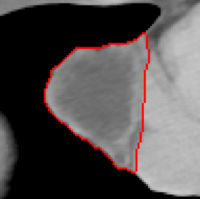}} \hfill 
 \fbox{\includegraphics[width=0.1\linewidth, height=0.1\linewidth]{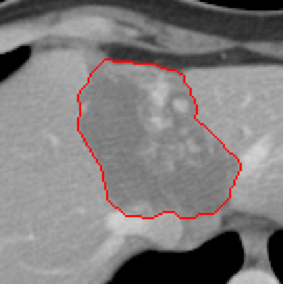}} \hfill
 \fbox{\includegraphics[width=0.1\linewidth, height=0.1\linewidth]{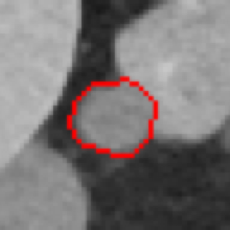}} \hfill
 \fbox{\includegraphics[width=0.1\linewidth, height=0.1\linewidth]{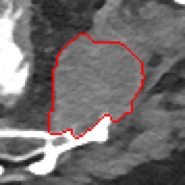}} \hfill
 \hspace{0.05\linewidth} \hfill
\end{center}
\vspace{-4mm}
\caption{Qualitative results. The $1^{st}$ row presents failed cases that rejected by radiologists. With the red curve delineate segmented lesion boundaries, the missed actual lesion region and the health tissue that be erroneously segmented are highlighted by yellow and blue arrows, respectively. The $2^{nd}$ row presents good cases. Better viewed in color.}
\label{fig:visual-example}
\end{figure*}

In Fig.~\ref{fig:curves}(a), we show the segmentation results on 2D CT slices arranged in the order of offsets with respect to the RECIST-slice. GrabCut with 3D RECIST estimation (GrabCut-3DE), which is generated from RECIST propagation, produces good segmentations ($\sim$91\%) on the RECIST-slice but degrades to 55\% mDICE when the offset rises to 4. This is mainly because 3D RECIST approximation often is not a robust estimation across slices. In contrast, the HNN trained with only RECIST slices, \ie{}, the model from Sec.~\ref{exp:cnn}, generalizes well with large slice offsets, achieving mean DICE scores of $>70$\% even when the offset distance ranges to 6. However, performance is further improved at higher slice offsets when using the proposed slice-propagated approach with 3 axial slices, \ie, WSSS-3, and even further when using slice-propagated learning with 5 and 7 axial slices, \ie, WSSS-5, and WSSS-7, respectively.  These results demonstrate the value of using our slice-propagated learning approach to generalize beyond 2D RECIST-slices into full 3D segmentation. Fig.~\ref{fig:curves}(b) further demonstrates the model improvements from slice-propagated learning using precision-recall curves. The categorized 2D and 3D segmentation results are tabulated in Table~\ref{tab:lesion-seg}. In addition, we display qualitative results as good and failed cases evaluated by a board-certificated radiologist in Fig.~\ref{fig:visual-example}.

\begin{table}[t!]
\begin{center}
\caption{Results of using different training masks, where GT refers to the manual segmentations. All results report mDICE $\pm$ std. GT results for the DeepLesion dataset are trained on the subset of $1,000$ annotated slices. See Sec.~\ref{exp:cnn} for method details.} 
\vspace{-3mm}
\label{tab:2d-convnet}
{\scriptsize
\begin{tabular}{>{~}m{2.7cm}>{\centering}m{1.7cm} >{\centering}m{1.7cm} >{\centering}m{0.1cm} >{\centering}m{1.7cm} >{\centering\arraybackslash}m{1.7cm}} 
\toprule
& \multicolumn{2}{c}{Lymph Node} & &\multicolumn{2}{c}{DeepLesion (on RECIST-Slice)}\\ \cmidrule{2-3} \cmidrule{5-6}
Method & CNN & CNN-GC & & CNN & CNN-GC \\ \midrule
UNet${}+{}$GT 				& \textBF{0.729$\pm$0.08} & 0.838$\pm$0.07& & 0.728$\pm$0.18 & 0.838$\pm$0.16  \\
HNN${}+{}$GT 				& 0.710$\pm$0.18 & \textBF{0.845$\pm$0.06} & & 0.837$\pm$0.16 & 0.909$\pm$0.10  \\
HNN${}+{}$RECIST-D         	& 0.614$\pm$0.17 & 0.844$\pm$0.06 & & 0.644$\pm$0.14 & 0.801$\pm$0.12 \\
HNN${}+{}$GrabCut-R       	& 0.702$\pm$0.17 & 0.844$\pm$0.06 & & \textBF{0.906$\pm$0.09} & \textBF{0.915$\pm$0.10}  \\
\bottomrule
\end{tabular}
}
\vspace{-6mm}
\end{center}
\end{table}


\begin{table*}[t!]
\begin{center}
\caption{Mean DICE scores for 
	lesion volumes. ``HNN'' is the HNN~\cite{xie_2015_hnn} trained on GrabCut-R from RECIST slices and ``WSSS-7'' is the proposed approach trained on 7 successive CT slices. See Sec.~\ref{exp:wsss} for method details.\vspace{-3mm}}
\label{tab:lesion-seg}
{\scriptsize
\begin{tabular}{>{~}m{2cm}>{\centering}m{1cm} >{\centering}m{1cm} >{\centering}m{1cm} >{\centering}m{1cm} >{\centering}m{1cm} >{\centering}m{1cm} >{\centering}m{1cm} >{\centering}m{1cm} >{\centering\arraybackslash}m{1cm}}
\toprule
Method & Bone & AB & MD & Liver & Lung & Kidney & ST & Pelvis & Mean \\ \midrule
GrabCut-3DE  & 0.654 		& 0.628  		& 0.693  		& 0.697  		& 0.667  		& 0.747  		& 0.726  		& 0.580 		& 0.675 \\
HNN     & 0.666 		& 0.766  		& 0.745  		& 0.768  		& 0.742  		& 0.777  		& \textBF{0.791}& \textBF{0.736}& 0.756 \\
WSSS-7    & \textBF{0.685}& 0.766  		& \textBF{0.776}& \textBF{0.773}& 0.757  		& \textBF{0.800}& 0.780  		& 0.728 		& 0.762 \\
WSSS-7-GC & 0.683 		& \textBF{0.774}& 0.771  		& 0.765  		& \textBF{0.773}& \textBF{0.800}& 0.787  		& 0.722 		& \textBF{0.764} \\ \bottomrule
\end{tabular}
\vspace{-6mm}
}
\end{center}
\end{table*}

\section{Conclusion}
We present a simple yet effective weakly supervised segmentation approach that converts massive amounts of RECIST-based lesion diameter measurements (retrospectively stored in hospitals' digital repositories) into full 3D lesion volume segmentation and measurements. Importantly, our approach does not require pre-existing RECIST measurement on processing new cases. The lesion segmentation results are validated quantitatively, \ie, 91.5\% mean DICE score on RECIST-slices and 76.4\% for lesion volumes. We demonstrate that our slice-propagated learning improves performance over state-of-the-art CNNs. Moreover, we demonstrate how leveraging the weakly supervised, but large-scale data, allows us to outperform fully-supervised approaches that can only be trained on subsets where full masks are available. Our work is potentially of high importance for automated and large-scale tumor volume measurement and management in the domain of precision quantitative radiology imaging.

\vspace{2mm}
{\bf Acknowledgments.} This research was supported by the Intramural Research Program of the National Institutes of Health Clinical Center and by the Ping An Insurance Company through a Cooperative Research and Development Agreement. We thank Nvidia for GPU card donation.

\bibliographystyle{splncs}
\bibliography{egbib}
\end{document}